\documentclass[12pt]{article}

\newcommand{\beginsupplement}{%
        \setcounter{table}{0}
        \renewcommand{\thetable}{S\arabic{table}}%
        \setcounter{figure}{0}
        \renewcommand{\thefigure}{S\arabic{figure}}%
     }
\usepackage{tabularx}
\usepackage{color,soul}
\usepackage{upgreek}
\usepackage{mathastext}
\usepackage{setspace}

\usepackage{multirow}
\usepackage[labelsep=endash]{caption}
\usepackage{booktabs}
\usepackage{longtable}
\usepackage{scicite}
\usepackage{graphicx}
\usepackage{times}
\usepackage{color,soul}
\usepackage{upgreek}
\usepackage{mathastext}
\usepackage{setspace}
\usepackage{eqnarray,amsmath}
\usepackage{algpseudocode}
\usepackage{algorithm,algcompatible}
\algnewcommand\INPUT{\item[\textbf{Input:}]}%
\algnewcommand\OUTPUT{\item[\textbf{Output:}]}%

\usepackage{ragged2e}

\newenvironment{sciabstract}{%
\begin{quote} \bf}
{\end{quote}}

\title{Backpropagation-free Training of Deep Physical Neural Networks}

\author
{Ali Momeni,$^{1}$ Babak Rahmani,$^{2}$ Matthieu Malléjac,$^{1}$ Philipp del Hougne,$^{3}$ \\Romain Fleury$^{1\ast}$\\
\\
\footnotesize{$^{1}$Laboratory of Wave Engineering, Department of Electrical Engineering, EPFL, CH-1015 Lausanne, Switzerland,}\\
\footnotesize{$^{2}$now at Microsoft Research, 198 Cambridge Science Park, CB4 0AB Cambridge, UK,}\\
\footnotesize{$^{3}$Univ Rennes, CNRS, IETR - UMR 6164, F-35000, Rennes, France.}\\
\footnotesize{$^\ast$   romain.fleury@epfl.ch}
}

\date{}

\begin{document} 


\baselineskip24pt


\maketitle


\begin{sciabstract}
  Recent years have witnessed the outstanding success of deep learning in various fields such as vision and natural language processing. This success is largely indebted to the massive size of deep learning models that is expected to increase unceasingly. This growth of the deep learning models is accompanied by issues related to their considerable energy consumption, both during the training and inference phases, as well as their scalability. Although a number of work based on unconventional physical systems, such as wave-based frameworks, addresses the issue of energy efficiency in the inference phase, efficient training of deep learning models has remained elusive. So far, training of digital deep learning models mainly relies on backpropagation, which is not suitable for physical implementation as it requires perfect knowledge of the computation performed in the so called forward pass of the neural network. Here, we tackle this issue by proposing a simple deep neural network architecture augmented by a biologically plausible learning algorithm, referred to as "model-free forward-forward training". This new route enables direct training of deep physical neural networks consisting of layers of arbitrary physical nonlinear systems, without requiring detailed knowledge of the nonlinear physical layers' properties.  We show that our method outperforms state-of-the-art hardware-aware training methods by improving training speed, decreasing digital computations, and reducing power consumption in physical systems, particularly optics. We demonstrate the robustness and adaptability of the proposed method, even in systems exposed to dynamic or unpredictable external perturbations where all other hardware-aware schemes fail. To showcase the universality of our approach, we train diverse wave-based physical neural networks that vary in the underlying wave phenomenon and the type of non-linearity they use, to perform vowel and image classification tasks experimentally. This work paves the way for the ambitious goals of hybrid training massive physical neural networks, which can offer high-speed and lower energy consumption not only for inference but also during the training phase.
\end{sciabstract}


\section*{Introduction}
Deep learning has emerged as a breakthrough technology with outstanding success in various fields such as vision, natural language processing (NLP), and speech recognition\cite{lecun2015deep,deng2014deep}. Although these algorithms attempt to mimic the functioning of the human brain, they are essentially executed on a software level using traditional von Neumann computing hardware. Nevertheless, artificial neural networks (ANNs) based on digital computing are currently experiencing challenges concerning energy consumption and processing speed \cite{patterson2021carbon}. An example of the considerable energy usage involved in training language models can be seen in the case of GPT-3, which has 175 billion parameters and required 1.3 GWh of electricity during training \cite{asghari2011energy}. As a result, researchers are increasingly exploring the implementation of ANNs on alternative physical platforms, including but not limited to optics \cite{wetzstein2020inference,lin2018all,wu2020neuromorphic,hughes2019wave,momeni2022electromagnetic}, spintronic \cite{romera2018vowel,grollier2020neuromorphic}, nanoelectronic devices \cite{chen2020classification,ruiz2020deep,shen2017deep}, and photonic hardware \cite{wetzstein2020inference,appeltant2011information}, as well as acoustic systems \cite{momeni2023physics,momeni2022learning}.  Currently, two primary methods are mainly used. The first requires the designing of hardware that implements trained mathematical transformations through strict operation-by-operation mathematical isomorphism, which is mainly aimed at the inference phase of deep learning \cite{burr2017neuromorphic,markovic2020physics,prezioso2015training,anderson2023optical}. The second category, known as deep Physical Neural Networks (PNNs), involves training the physical transformations of the hardware directly to execute the desired computations.
PNNs offer the potential to create more scalable, energy-efficient, and faster neural network hardware by leveraging physical transformations and foregoing the conventional software–hardware division \cite{wright2022deep,nakajima2022physical}.

\begin{figure}[h!]
	\centering
	\includegraphics[width=15cm]{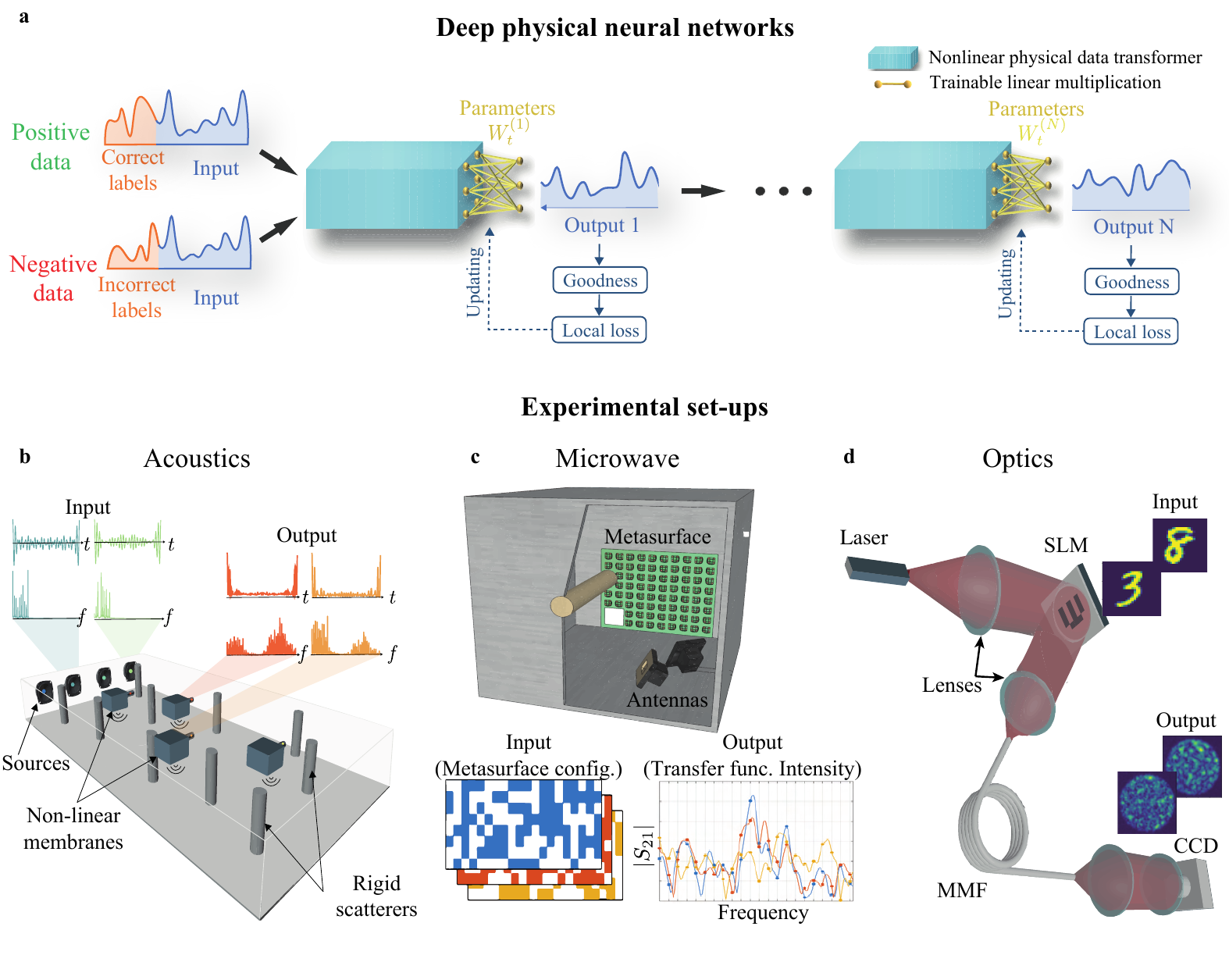}
	\caption{\textbf{Deep physical neural networks. } \textbf{a}, A simple and physics-compatible deep neural network that employs a sequence of nonlinear physical data transformers augmented by trainable matrix multiplications, trained by the model-free forward-forward (MF-FF) technique. At each layer, the nonlinear physical data transformer executes nonlinear mapping between the input and output space, and the MF-FF trains the trainable linear multiplications to determine the optimal decision boundary for positive and negative data. We consider three physical systems that vary in terms of the underlying wave phenomenon and the type of non-linearity.   \textbf{b}, In acoustics, input data is encoded into the intensity of sound waves at different frequencies injected on the left side of the cavity. Sound waves propagate through a chaotic cavity that comprises multiple rigid cylindrical diffusers and nonlinear membranes. The transformed waveforms are received by multiple microphones.  \textbf{c},   In the chaotic microwave cavity, input data is  encoded into the programmable metasurface configuration inside the metallic disordered cavity. The outputs are obtained from the
     waves’ spectra (transfer function).   \textbf{d}, In optical setup, input data is encoded onto the SLM, and after passing through the MMF, the resulting optical intensity is measured on the CCD camera.
}\label{Fig_1}
\end{figure}
In a similar fashion to digital deep learning, increasing the depth of physical networks (deep PNNs) holds great potential for enhancing performance as it leads to an exponential expansion in network expression capabilities \cite{montufar2014number}. So far, the training of PNNs has predominantly relied on a method called backpropagation (BP), which has been highly successful in the training of the digital ANNs \cite{lillicrap2020backpropagation}. Yet, there are several reasons why BP is not a suitable choice for PNNs, one of which is the complexity and lack of scalability in the physical implementations of BP operations  in the hardware \cite{pai2022experimentally,guo2021backpropagation,hughes2018training,bandyopadhyay2022single}. For example, in \cite{pai2022experimentally}, the authors use the adjoint variable method to experimentally implement BP in photonic neural networks which is limited to lossless and reciprocal PNNs. Additionally, their method requires three rounds of light propagation, either from left to right or right to left, throughout the PNN, which remains unachievable in the majority of PNN setups.   Most proposals for PNNs implement BP calculations on an external computer using a digital twin of the physical system, commonly referred to as in-silico training. This usually comes at the cost of speed and an increase in the energy consumption during training. Additionally, the model might not accurately represent the real physical system which can lead to a potential simulation–reality gap and unfaithful inference time prediction \cite{wu2020neuromorphic,ruiz2020deep,romera2018vowel,shen2017deep,prezioso2015training,hughes2019wave,lin2018all,miller2014evolution,bueno2018reinforcement}. Recent work has made some progress towards addressing these issues. The physics-aware training method based on BP (PA-BP) \cite{wright2022deep} is the current state-of-the-art training framework that circumvents some problems of the in-silico methods. However, the Achilles heel of PA-BP is that it still requires a differentiable digital model for the backward pass. This imposes some limitations on the use cases of the PA-BP, such as the slow training speed, and the high power consumption due to the digital backward pass and the need for extra memory accommodating the backward model of the PNN on a digital computer. Furthermore, as becomes evident later, in the event that the physical system undergoes strong perturbations, reusing PA-BP-trained PNNs, or even their fine-tuned versions could be quite challenging and in some cases futile, necessitating training of the models from scratch.

Another significant drawback of BP is its reliance on having complete knowledge of the computations graph carried out during the forward pass to accurately compute derivatives \cite{hinton2022forward,nakajima2022physical,zhu2022contrastive,lee2023symba,srinivasan2023forward}. When a black box is inserted in the forward pass, BP becomes impossible.  Therefore, researchers are seeking alternative training methods for PNNs. For example, an approach that has been recently explored for training physical networks is the augmented Direct Feedback Alignment (DFA) method \cite{nakajima2022physical}, which aims to avoid the need for a differentiable digital model. However, this method is only compatible with certain physical networks where the nonlinear and linear layers can be separated. Furthermore, determining the nonlinearity form for PNNs through optimization procedures is still an ongoing challenge. 


Here, we propose a simple and physics-compatible PNNs architecture augmented by a biologically plausible learning algorithm, called the model-free forward-forward (MF-FF) training. The proposed method enables direct training  of arbitrary PNNs locally without the need to know the nonlinear physical layers and train a digital twin model. To the best of our knowledge this is the first work demonstrating a BP-free training of PNNs that circumvents all challenges associated with the hardware-realization of BP.
In this BP-free contrastive learning method, the standard backward pass, typically performed by a digital computer, is replaced with a single forward pass through a physical system. This substitution can significantly improve training speed compared to other hardware aware training framework, reduce digital computations and memory usage, and lower power consumption in the training phase of wave-based PNNs.
In this work, we benchmark our learning framework against the state-of-the-art physics-aware training schemes proposed in the literature, i.e., in-silico and backpropagation methods. We show the robustness and adaptability of the proposed method compared to its sister schemes, even in systems exposed to unpredictable external perturbations. 
To showcase the universality of our approach, we perform experimental vowel and image classification using three wave-based systems that differ in terms of the underlying wave phenomenon and the type of non-linearity involved. Our first example consists of a chaotic acoustic cavity implemented with non-linear scatterers. The second example is a chaotic microwave cavity whose transfer function is massively parameterized by a programmable metasurface with structural non-linearity. Our third example is an optical multimodal fiber with readout non-linearity .
Our approach results in high-accuracy achieving hierarchical classifiers that make use of the distinct physical transformations of each system and offers a pathway to significantly enhance the energy efficiency and speed of deep learning not only during inference but also during the training phase.

\section*{Model-free forward-forward training}
Figure~\ref{Fig_1}a shows a simple and physics-compatible deep PNN including  $\textit{N}$ nonlinear physical data transformers augmented by  trainable linear multiplications.  Each nonlinear physical data transformer performs a nonlinear mapping between the input and output space “effortlessly”, followed by the use of augmented trainable linear multiplication to classify distinct classes via a local and contrastive training algorithm.  The orientation of neuron activities, the output of each layer,  is passed to the next layer. The subsequent layer then carries out the same process hierarchically on the output of the previous layer.
 This architecture shares some similarities with conventional deep Reservoir Computing (deep-RC) systems \cite{gallicchio2017deep}. However, here all augmented linear multiplications are trained, in contrast to the traditional deep-RC where only the final layer is trained\cite{rafayelyan2020large,dong2019optical}.
If nonlinear physical transformations possess some form of memory, the network can be categorized as a trainable deep-RC.  
 Among the various contrastive learning approaches \cite{srinivasan2023forward}, the forward-forward algorithm \cite{hinton2022forward}, inspired by Boltzmann machines \cite{hinton1986learning}, was recently proposed by Geoffrey Hinton   and has undergone further improvement through several proposals in computer science \cite{zhao2023cascaded,ororbia2023predictive}. Here, we implement the model-free version of the forward-forward algorithm, for the aforementioned PNNs-based architecture.
 \begin{figure*}[t]
	\includegraphics[width=1\textwidth]{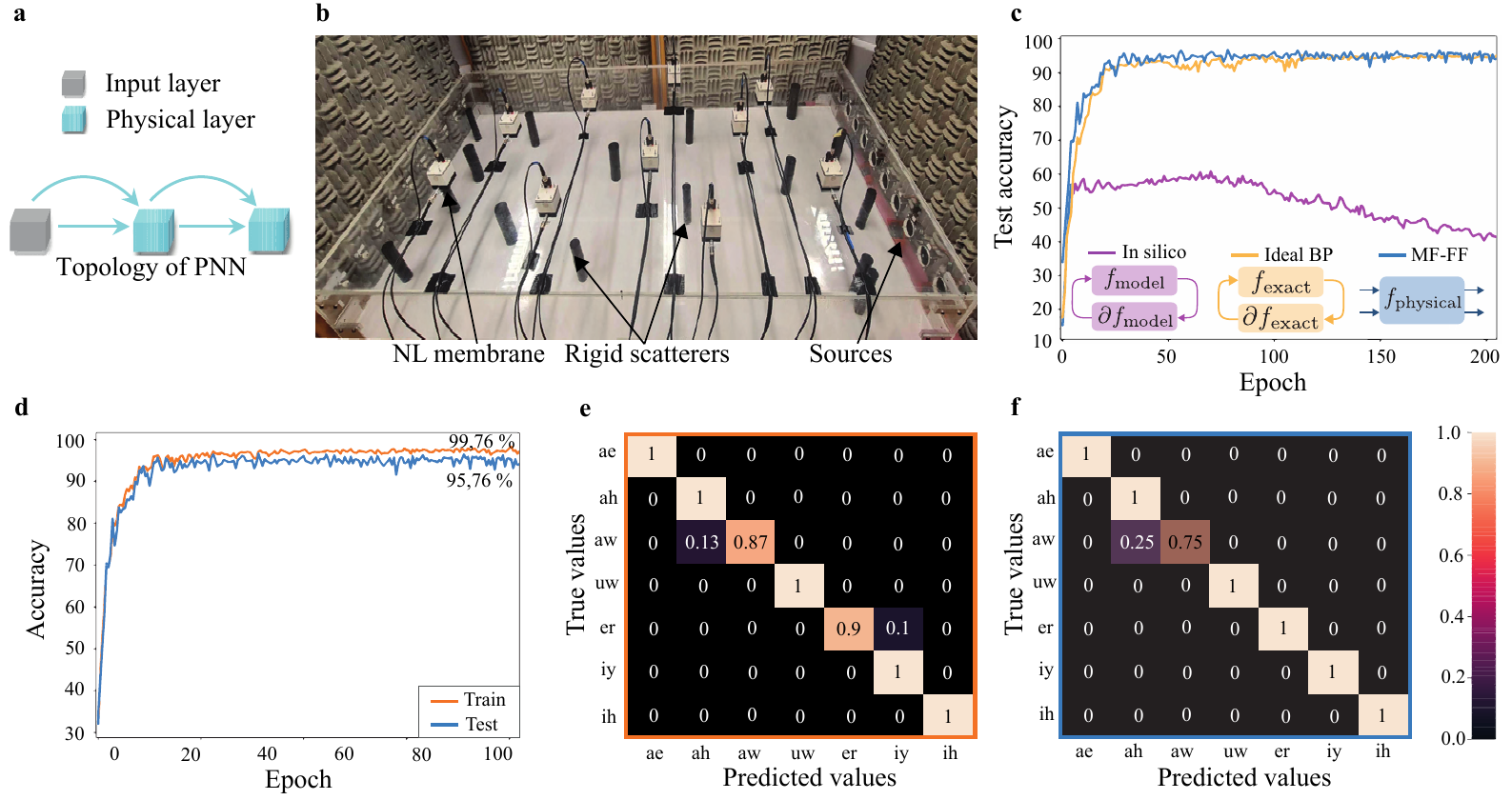}
	\centering
	\caption{\textbf{Acoustic-PNN.} \textbf{a}, The topology of acoustic-PNN consists of a two-layer PNN with skip connections. Each layer comprises an acoustic-PNN augmented by trainable matrix multiplication.    \textbf{b}, Photograph of the experimental setup. \textbf{c}, Comparison of test accuracy  versus training epoch with in silico, ideal back-propagation, and MF-FF algorithm for the vowel recognition task.  \textbf{d}, The train and test classification accuracy versus training epoch for the vowel recognition task.  \textbf{e} and  \textbf{f}, The confusion matrix for the PNN on the train and test sets, respectively.  }\label{Fig_2}
\end{figure*}

Each nonlinear physical system performs a nonlinear transformation on input data, which can be expressed as $ \it {{h^{(l)}} = f_{\text{N}}^{(l)}(W_{\text{p}}^{(l)}x^{(l)})}$, where $\it x^{(l)}$,  $ \it W_{\text{p}}^{(l)}$, and $ \it f_{\text{N}}^{(l)}$ correspond to the physical inputs (e.g., optical intensity, electric voltage, vibration),  physical interconnections
(e.g., optical, electrical, or mechanical coupling) in the physical
system, and physical nonlinearity (e.g., nonlinear optical, magnetic, or mechanical effects) in layer $\it l$, respectively. 
Indeed, $ \it W_{\text{p}}^{(l)}$ and $ \it f_{\text{N}}^{(l)}$ signify  mixing operation and non-linear kernel of the $\it l-$th  physical systems. Afterward, the output of layer $\it l$ can be expressed as the multiplication of $\it {h^{(l)}}$ by the augmented trainable weight matrix $\it {W_{\text{t}}^{(l)}}$, $\it y^{(l)}=W_{\text{t}}^{(l)}{h^{(l)}}$.
Such trainable matrix multiplications can be performed through either digital or physical systems, for instance using Mach-Zehnder Interferometer (MZI) integrated photonics \cite{zhou2022photonic} Spatial Light Modulators (SLMs) in optics\cite{anderson2023optical,matthes2019optical}. The goal here is to train $W_{\text{t}}^{(l)}$ locally. 
Instead of a forward and backward pass, here, we use two physical forward passes: a positive and a negative forward pass through the physical systems, each running on different physical inputs. The positive physical pass, $\it y_{\text{pos}}^{(l)}=W_{\text{t}}^{(l)} f_{\text{N}}^{(l)}(W_{\text{p}}^{(l)}x_{\text{pos}}^{(l)})$, uses positive inputs that include the input dataset and correct labels, while the negative physical pass, $\it y_{\text{neg}}^{(l)}=W_{\text{t}}^{(l)} f_{\text{N}}^{(l)}(W_{\text{p}}^{(l)}x_{\text{neg}}^{(l)})$, uses negative inputs that include the  input dataset and incorrect labels (see \textcolor{blue}{Fig. \ref{Fig_1}a}).
In each layer, we calculate the so-called "goodness" function, defined as the sum of the squared activities for the positive and negative physical passes. For instance, in physical systems such as optics, the squared activities correspond to the optical intensity on the CCD camera. 
Eventually, for each layer $l$, $W_{\text{t}}^{(l)}$  is trained by minimizing the following loss function
\begin{equation}
\it
L^{(l)}= log \bigg(1+exp \bigg({-\theta \big(\sum_{j}{ y_{\text{pos},j}^{(l)}}^2 -\sum_{j}{ y_{\text{neg},j}^{(l)}}^2 \big) \bigg)}\bigg)
\end{equation}
where $\theta$ is a scale factor.  During the inference phase, we input a particular label into the PNNs and accumulate the goodness values for all layers. This process is repeated for each label separately. The label with the highest accumulated goodness value is then selected as the output (see supplementary section S2 for more details).
The proposed method is also capable of integrating non-differentiable physical systems or components between the layers.

\section*{Diverse PNNs for vowel and image classification}

In \textcolor{blue}{Figure~\ref{Fig_1}}, we present three deep PNN classifiers for the vowel and Mnist (Modified National Institute of Standards and Technology database) handwritten digit classification task, based on three distinct physical systems. 
Although there have been proposals that explore wave-based analog computing for linear operations such as multiplication and convolution \cite{silva2014performing,xu202111,babaee2021parallel,wang2022optical,rajabalipanah2022parallel,del2018leveraging,momeni2021reciprocal,zhou2022photonic,momeni2021asymmetric}, it is important to note that PNNs require nonlinearity to effectively handle a wide range of learning tasks. These tasks include regression and classification, which require nonlinear operations for accurate performance. 
We consider three wave-based physical systems, each of which features a unique origin of physical nonlinearity and underlying wave phenomenon, highlighting the diversity of physical networks that can be employed.
We take advantage of the multiple scattering induced by the disordered environment in these physical systems (chaotic cavities and multimodal fibers) to leverage the mixing process. In addition to provide the required linear mixing, it also helps to intensify the overall nonlinearity \cite{momeni2023physics}.

\subsubsection*{Acoustic Chaotic Cavity with Non-Linear Scatterers}
In acoustics, an air-filled multimode cavity composed of multiple nonlinear meta-scatterers randomly placed on the cavity top wall and  multiple rigid scatterers inside the cavity is employed (materials and methods, supplementary section S3). The nonlinear meta-scatterers are designed based on an active control strategy.  Herein, the positive and negative data are encoded onto the amplitude of each frequency component composing the excitation waveforms, that are then injected into the nonlinear system through loudspeakers positioned on the right side of the cavity.  The output of the physical system is measured using microphones below the metascatterers. 
We investigate the vowel classification performance of two layers acoustic-PNN (see \textcolor{blue}{Fig.~\ref{Fig_2}a}). In order to compare the results of MF-FF with ideal BP, and in-silico training, we accurately model the forward pass of acoustic-PPN by a digital neural network (see the supplementary section S3). When the acoustic-PNN is trained using MF-FF,  it performs the classification task with 97.62\% and 94.23\% train and test accuracy, respectively (see \textcolor{blue}{Figs. \ref{Fig_2}d-f}).  \textcolor{blue}{Figure~\ref{Fig_2}c} shows the comparison of the classification results obtained for MF-FF, ideal BP, and in-silico training.
A schematic visual representation of the aforementioned methods is provided above \textcolor{blue}{Fig. \ref{Fig_2}c}. The complete comparison between different hardware-based training methods is provided in supplementary section S2.

As evidenced by \textcolor{blue}{Fig.~\ref{Fig_2}c}, in-silico training performs poorly, reaching only a maximum vowel classification accuracy of $\sim$60\%. When there is a gap between the reality and the simulation of a physical system ( called the reality-simulation gap), the accuracy of inference will decrease. In contrast, MF-FF succeeds in accurately training the acoustic-PNN, performing similarly to the ideal BP algorithm used as a baseline. The key advantage of MF-FF stems from the execution of both forward passes through the physical hardware, rather than simulations.

\begin{figure}[t!]
	\centering
	\includegraphics[width=\textwidth]{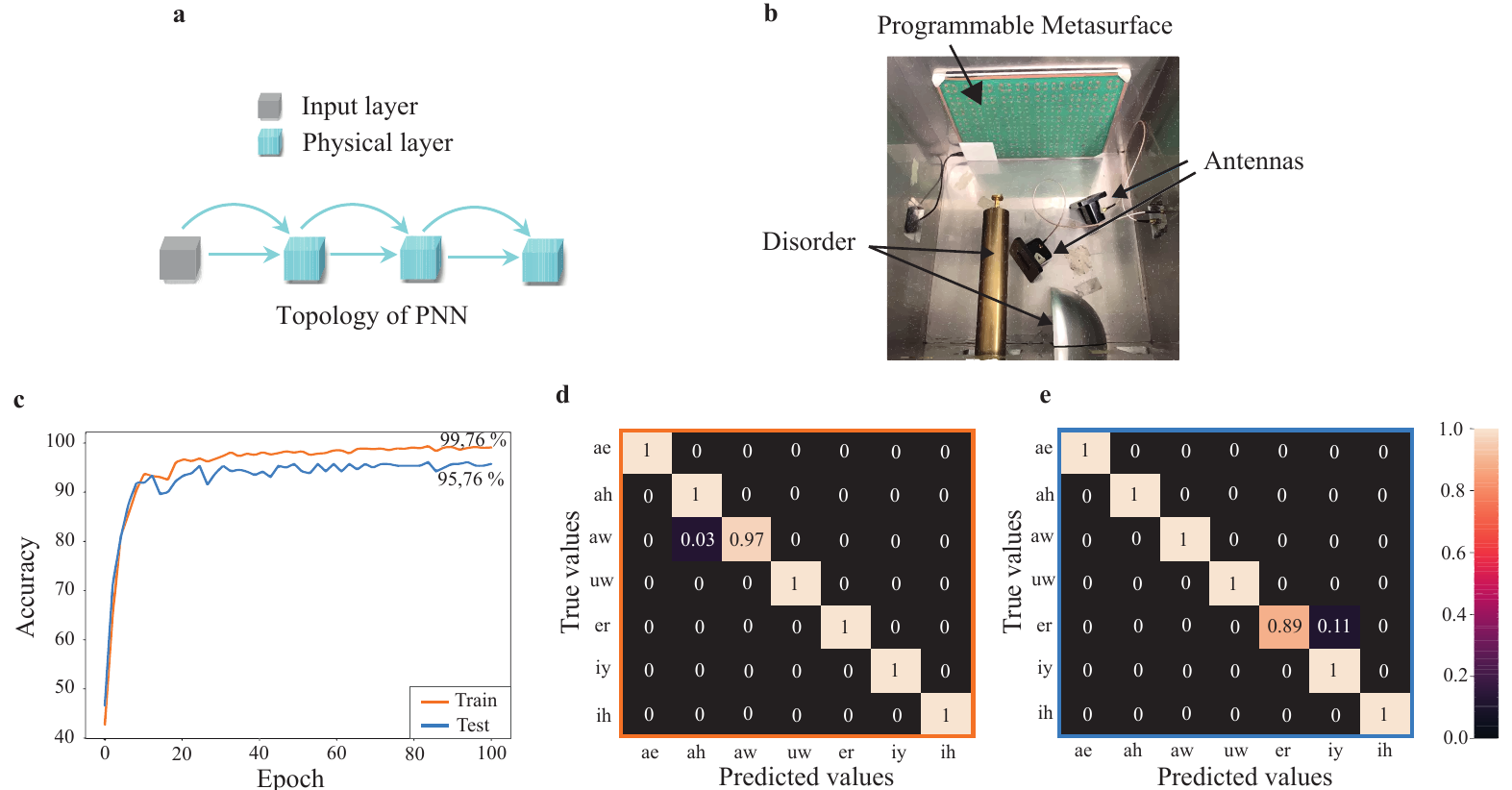}
	\caption{\textbf{ Microwave-PNN.} \textbf{a}, The topology of microwave-PNN consists of a three-layer PNN with skip connections. Each layer comprises a microwave-PNN augmented by trainable matrix multiplication.    \textbf{b}, Photograph of the experimental setup. \textbf{c}, The train and test classification accuracy versus training epoch for the vowel recognition task.  \textbf{d} and  \textbf{e}, The confusion matrix for the PNN on the train and test set, respectively. }\label{Fig_3}
\end{figure}

\subsubsection*{Microwave Massively Parametrized Chaotic Cavity with Structural Non-Linearity}

In the microwave regime, we leverage a ``structural non-linearity'' such that we can implement nonlinear mathematical operations at low power levels with a linear scattering system. Our starting point is an irregularly shaped electrically large metallic enclosure with strong modal overlap. It is coupled via two coax-to-waveguide adapters to two asymptotic scattering channels, and the transfer function between these two channels can be measured using standard equipment such as a vector network analyzer. We then massively parametrize this cavity by covering one of its walls with a programmable metasurface. For each meta-atom and each polarization, we can individually configure the local boundary conditions of the cavity with 1-bit precision (two possible states). Our setup is shown in Fig.~\ref{Fig_3}(b) and very similar to one that was recently used to implement with high fidelity and in situ reprogrammability desired \textit{linear} transfer functions for signal differentiation~\cite{sol2022meta} and routing~\cite{sol2023reflectionless}. In the present work, however, we seek a \textit{non-linear} mapping. Hence, we define the metasurface configuration as the  input and the transfer function as the output of our mathematical operation. Indeed, this relation is in general nonlinear due to the mutual coupling between meta-atoms caused by their proximity and, more importantly, the reverberation~\cite{rabault2023tacit}. While previous work in Ref.\cite{del2018leveraging} sought to limit the reverberation in order to implement a linear transformation with the metasurface configuration as input and the transfer function as output, here we deliberately seek to maximize the reverberation to boost the nonlinearity (further explained in the supplementary section S4).
Incidentally, this type of reverberation-induced structural nonlinearity was recently also transposed to the optical domain\cite{eliezer2022exploiting}.  
  \begin{figure}[t!]
	\centering
	\includegraphics[width=\textwidth]{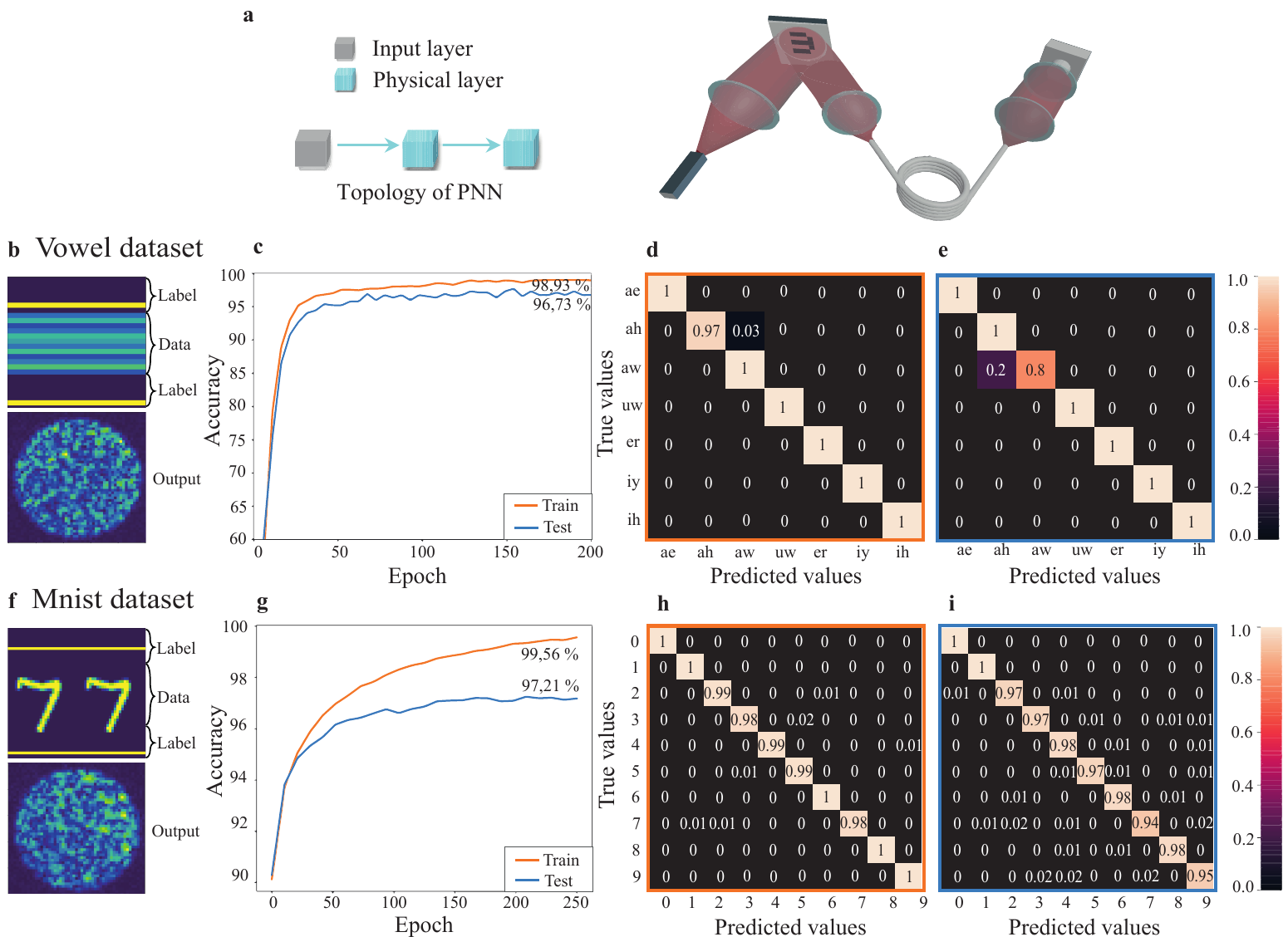}
	\caption{\textbf{ Optics-PNN.} \textbf{a}, The topology of optics-PNN consists of a two-layer PNN. Each layer comprises an optics-PNN augmented by trainable matrix multiplication.    \textbf{b} and \textbf{f}, An example of input data, including raw data and its label representation on SLM, along with its corresponding output on a CCD camera for vowel and Mnist datasets, respectively. \textbf{c} and \textbf{g}, The train and test classification accuracy versus training epoch for the vowel recognition and Mnist tasks, respectively.  \textbf{d} and  \textbf{e}, The confusion matrix for the PNN on the train and test set for the vowel recognition task, respectively. \textbf{h} and  \textbf{i}, similar results for the Mnist task. }\label{Fig_4}
\end{figure}

We randomly group our programmable metasurface's 152 degrees of freedom into 40 macropixels since our mathematical operation requires 40 inputs. We define our mathematical operation's outputs as the transfer function intensities at {twenty} decorrelated frequencies within the bandwidth of operation of the programmable metasurface (400~MHz around 5.2~GHz). Note that, in addition to the structural non-linearity, we hence add a readout non-linearity by working with the transfer function's intensity. In order to flexibly evaluate the proposed approach, based on 50000 experimentally measured pairs of a random metasurface configuration and the corresponding transfer function, we learn a digital surrogate forward model of the configuration to transfer function intensity mapping (see supplementary section S4). Then, we construct the three-layer microwave-PNN shown in \textcolor{blue}{Fig.~\ref{Fig_3}a} and train it according to the MF-FF. The training converges after roughly 20 epochs and the achieved classification accuracy on unseen test data reaches around 96~\% (see \textcolor{blue}{Figs.~\ref{Fig_3}c-e}).

\subsubsection*{Optical Multimode Fiber with Readout Non-Linearity}
In the optics part, we use an optical system that comprises an SLM, a scattering medium consisting of a step-index multimode fiber (MMF), and a CMOS camera (see \textcolor{blue}{Fig.~\ref{Fig_1}d}, materials and methods). In this setup, the positive and negative datasets are encoded onto the SLM, and after passing through the MMF, the resulting optical intensity is measured on the camera. The physical optical system performs a complex spatiotemporal transformation.  
This transformation involves the propagation of spatially modulated laser pulses through an MMF. The propagation of an ultrashort pulse inside a MMF is a highly complex process that involves spatial and temporal interactions of electromagnetic waves coupled to hundreds of different propagation modes \cite{teugin2021scalable,oguz2022programming}. Although this transformation is linear in the complex domain, the process becomes nonlinear due to the data being encoded onto the phase (SLM) and the subsequent measurement of the intensity squared on the camera.  Here, we use a two-layer optics-PNN to perform classification tasks on two different datasets: Vowel and Mnist (see \textcolor{blue}{Fig. \ref{Fig_4}} and supplementary section S5 for further details).
The optics-PNN achieved an impressive  classification performance on both vowel and Mnist datasets. We obtain 98.93\% and 96.73\% accuracy on the training and test vowel datasets, respectively. Using only two-layer optics-PNN, the model achieves 99.56\% and 97.21\% accuracy on the training and test Mnist datasets (see corresponding results in \textcolor{blue}{Figs. \ref{Fig_4}c-h}).
These results demonstrate the ability to transition from an expensive digital processor to a fast, energy-efficient hybrid physical-digital processor, showcasing the potential for optimizing both performance and energy efficiency in machine learning applications. 

\textcolor{black}{
After posting the initial version of this work on arXiv, we became aware that a related idea and experiment showed for optics was being pursued by another team, see preprint at \cite{oguz2023forward}.}

\begin{figure}[t!]
	\centering
	\includegraphics[width=\textwidth]{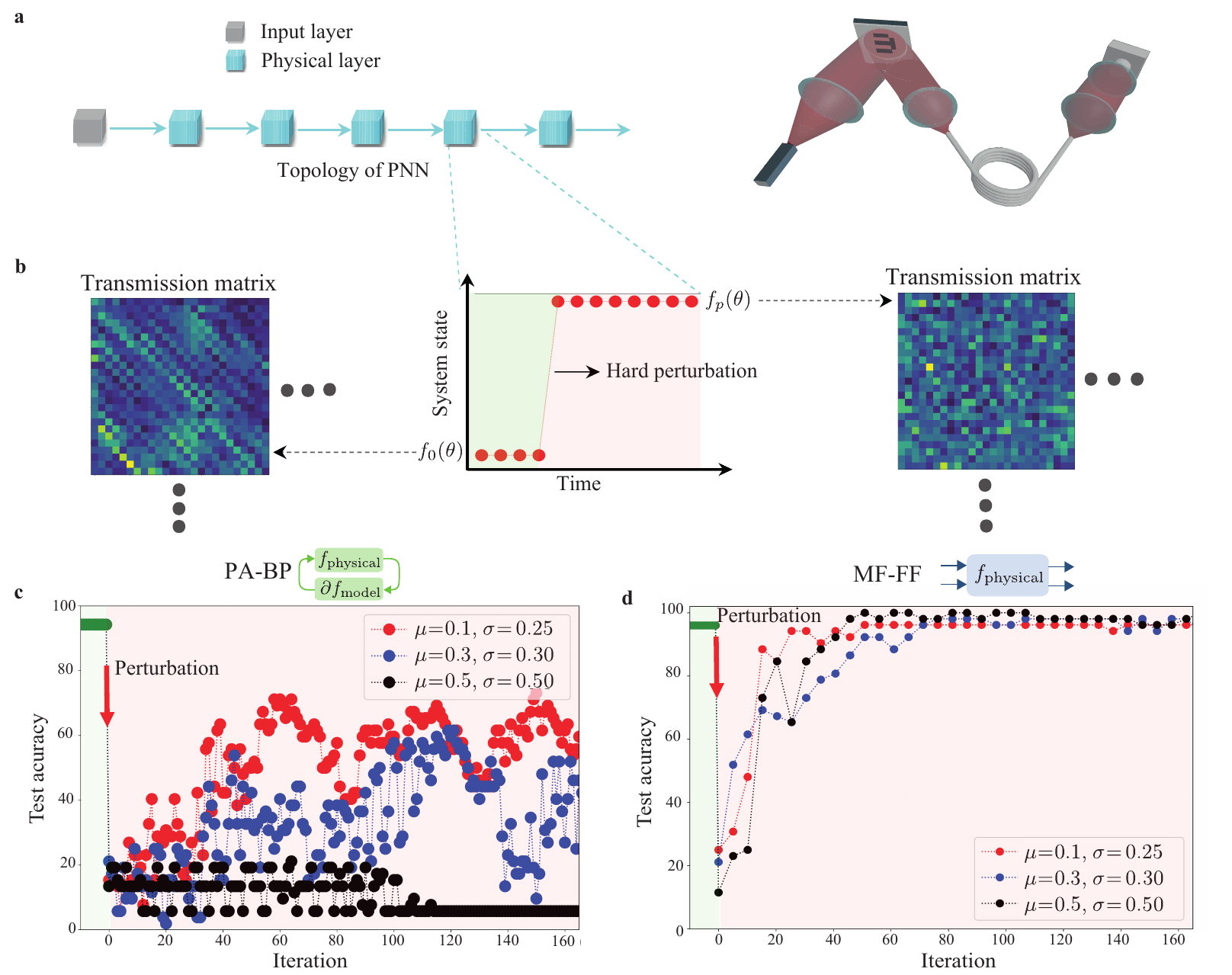}
	\caption{\textbf{ Robustness of deep PNNs against  unpredictable external perturbations.} \textbf{a}, A deep PNN consists of six layers of optics-PNN  augmented by trainable matrix multiplication. The deep PNNs trained on vowel datasets and is currently in the inference phase   \textbf{b}, Applying hard-perturbation by adding Gaussian noise with the mean of $\mu$ and standard deviation of $\sigma$ to the transmission matrix of MMF. \textbf{c} and \textbf{d}, A comparison between PA-BP \cite{wright2022deep} and the proposed MF-FF method is presented, with the focus on their ability to recover the classification accuracy after applying perturbation.    }\label{Fig_5}
\end{figure}

\section*{Real-Time Adaptable Learning}
We now aim to show the superior robustness of MF-FF in the context of real-time and adaptable learning, where the physical data transformer may undergo changes due to slow dynamics of the physical system during the runtime or external hard perturbations. 
Let us consider a deep optics-PNN with six layers, as depicted in \textcolor{blue}{Fig.~\ref{Fig_5}a}, which has already been trained on vowel datasets and is currently in the inference phase. The transformation function of each physical system is $f_0(\theta)$, where $\theta$ is the physical input. We perturb the physical systems at a specific time (examples of such perturbations include changes in the MMF state or the positions of lenses or masks, etc), which results in a change in the transformation function of each physical system from $f_0(\theta)$ to $f_p(\theta)$ (\textcolor{blue}{ see Fig.~\ref{Fig_5}b}). We simply perturb the transmission matrix of the optical setup by adding Gaussian noise with mean $\mu$ and standard deviation $\sigma$ to simulate this situation.  As we observe in \textcolor{blue}{ Fig.~\ref{Fig_5}d}, the test accuracy drops as expected after applying the perturbation. The question now is whether the training method can restore the accuracy by retraining the optics-PNN after some epochs. We compare our results with the PA-BP method \cite{wright2022deep} which uses a digital model for the backward pass and the physical system for the forward pass. As shown in  \textcolor{blue}{ Fig.~\ref{Fig_5}c}, PA-BP struggles to restore accuracy with increasing perturbation intensity. For instance, the test accuracy oscillates around 55\% for a small perturbation (red dots in \textcolor{blue}{ Fig.~\ref{Fig_5}c}) and worsens further for more intense perturbations. In stark contrast, the proposed MF-FF can easily recover accuracy after a few epochs, regardless of the intensity of the perturbation applied. (see \textcolor{blue}{ Fig.~\ref{Fig_5}d}). 

This striking adaptability owed to the fact that MF-FF executes both forward passes through the physical hardware, rather than digital models. In contrast, the PA-BP method uses a digital model that becomes completely inaccurate after hard-perturbation, necessitating re-training with a new dataset and wasting energy.
This study highlights the robustness and adaptability of the proposed model, even in dynamic or unpredictable environments.

\section*{Discussion}

Due to the unprecedented growth in the size of ANNs, including models such as GPT 4 with around hundreds of billion parameters, the cost of both the training and inference phases of these networks has increased exponentially. Training of such massive ANNs is particularly extremely expensive and time-consuming, with training times taking up to several months. Specialized hardware such as PNNs have the potential to drastically decrease these costs. 
A few methods have been proposed for training PNNs, but they all face issues such as mismatch between the forward model and the physical system or robustness issues. This is because these methods perform the entire backward pass through a digital computer during training, involving either a digital model in PA-BP or numerical simulations in in-silico training, which can hinder their effectiveness in the training phase.
MF-FF enables forward passes through physical systems, resulting in a significant speed-up not only during inference but also in the training phase. Additionally, one prominent advantage of MF-FF training is its robustness to an external perturbation. We observed that the MF-FF training is superior not only to the in-silico training but also to the current state-of-the-art PA-BP method. The performance gap between our method and the PA-BP continues to widen as the depth of the NN increases. This is because in PA-BP training, the mismatch between the gradients of the non-perturbed and perturbed systems accumulate over the depth of the ANN. Therefore, the more layers a network has, the more sensitive it is to an external  perturbation. This is in contrast to the MF-FF trained framework in which each layer of the network is trained separately without using a model or direct dependence to other layers that is required for BP.

Reference \cite{anderson2023optical}  has recently revealed the exceptional efficiency of optical transformer models during the "inference" phase (achieving a $>8,000 \times$ energy-efficiency advantage over state-of-the-art digital-electronic processors ). The training method proposed in this paper represents a viable candidate for "training" such optical transformer models, enabling the potential to achieve energy-efficiency and speed advantages. 
We further study the potential of PNNs in terms of reducing energy consumption and improving the computing speed compared to digital training/inference schemes of ANNs in Section S6 of the supplementary material. 

We conclude that even though PNNs have the potential to improve deep neural networks' speed and energy consumption to a large extent, a number of challenges still need to be addressed before PNNs can replace their digital counterparts. For example, it is not yet clear how some widely used mathematical operations, such as normalization units, including layer/batch normalization, could be implemented in hardware. Second, how the physical systems could be compactly scaled to billions of parameters. 
Therefore, we can expect, at least in short term, that PNNs will enable efficient hybrid combinations of in-hardware training/inference rather than a complete replacement for digital processing units.





\section*{Acknowledgments}
A.M. and R.F. acknowledge funding from the Swiss National Science Foundation under the Eccellenza grant number 181232.
P.d.H. and R.F. acknowledge funding from the ANR-SNF PRCI program (project ``MetaLearn''; ANR-22-CE93-0010-01). B.R. acknowledges that all materials pertaining to the optical experiment presented in this manuscript have been sourced from previously published material that is publicly available and was acquired during his tenure at the Laboratory of Applied Photonics Devices at EPFL. This work has no connection to his current employer in any capacity or form. A.M. and M.M. thank Dr. X. Guo for useful discussions.

\section*{Contributions}
A.M. conceived the idea, designed the computational engine, and carried out both the theoretical and numerical simulations as well as a part of the acoustic experiment. B.R. provided the optics data and interpretation of machine learning results. M.M. carried out the acoustic experiment. P.d.H. carried out the microwave experiment. R.F. supervised the project. All authors contributed to the interpretation of the results and the writing of the manuscript.
\section*{Supplementary materials}
Materials and Methods\\
Supplementary Text\\
Figs. S1 to S7\\
Table S1 and S2

\bibliography{scibib}

\bibliographystyle{Science}



\clearpage
\newpage

\beginsupplement
{ \centering \LARGE \textbf{Supplementary Materials for:}\\\begin{center} Backpropagation-free Training of Deep Physical Neural Network \end{center}} 
\author
{\centering Ali Momeni, Babak Rahmani, Matthieu Malléjac, Philipp del Hougne, \\Romain Fleury\\
\footnotesize{ \centering Corresponding author:   Romain.fleury@epfl.ch}
}

\baselineskip24pt

\textbf{This PDF file includes:}

\textbf{Section S1. Materials and Methods}

\;\;\;\;\;\;S1-1. Acoustic system

\;\;\;\;\;\;S1-2. Microwave system

\;\;\;\;\;\;S1-3. Optical system

\textbf{Section S2. Comparison of different PNN training algorithms}

\textbf{Section S3. Acoustics-PNN}

\;\;\;\;\;\;S3-1. Experimental setup

\;\;\;\;\;\;S3-2. Input-output transformation characterization    

\textbf{Section S4. Microwave-PNN}

\;\;\;\;\;\;S4-1. Experimental setup

\;\;\;\;\;\;S4-2. Input-output transformation characterization 

\textbf{Section S5. Optics-PNN}

\;\;\;\;\;\;S5-1. Experimental setup

\;\;\;\;\;\;S5-2. Input-output transformation characterization

\textbf{Section S6. Energy consumption and computing rate  analysis}

\newpage
\section*{Section S1. Materials and Methods}
\subsection*{S1-1. Acoustic system}
The acoustic system consists of an air-filled cavity 2 m long, 1 m wide and 0.2 m thick supporting 11 propagating modes below 500 Hz. The sealed parallelepiped is instrumented on one side with 10 loudspeakers (Monacor SP6-8SQ) delivering the input waveforms to the system, multiple rigid cylindrical diffusers inside the cavity to increase disorder, and 10 nonlinear resonators (NLRs) randomly positioned on the top layer of the cavity. Each resonator is made up of an electrodynamic loudspeaker (Visator FRWS 5 SC), enclosed in a cavity of volume $V_c = 144$ m$^2$ and an ICP microphone (PCB 130F20, 1/4 inch) placed in front of the diaphragm. From the measured front pressure $p_f$, a real-time feedback loop assigns a given current $i(t)$ to each NL resonator according to the following control law $i(t) = G_{NL}|p_f(t)|^{\alpha_{NL}}$, where $G_{NL}$ and $\alpha_{NL}$ are two tunable parameters that produce the nonlinearity in the system. The active control is performed on an FPGA-based Speedgoat performance real-time target machine controlled by the xPC target environment of MATLAB/SIMULINK. 4 of the 10 sources are used for training. Each delivers a different linear waveform with a 10 frequencies content in the $350-500$ Hz range with randomly weighted amplitudes.  The output of the system consists of 4 highly non-linear pressure waveforms measured at the output of the active control at 4 different locations.  The measurement is performed 10,000 times with different amplitude weighting for each run and each input.  

\subsection*{S1-2. Microwave system}

Our microwave system is shown in Fig. 3b of the main paper and consists of an irregularly shaped electrically large metallic cavity ($0.385\ \mathrm{m} \times 0.422\ \mathrm{m} \times 0.405\ \mathrm{m}$; $0.0658\ \mathrm{m}^3$) whose scattering properties are massively parametrized through a programmable metasurface. 
Two waveguide-to-coax adapters (RA13PBZ012-BSMA-F) connect the cavity to two asymptotic scattering channels. The 2-bit programmable metasurface contains 76 meta-atoms that cover 8 \% of the cavity surface and efficiently modulate the field inside the cavity within a 400 MHz interval centered on 5.2 GHz. The two bits of control per meta-atom are assigned to the two orthogonal field polarizations. Using a vector network analyzer (Rhode \& Schwarz ZVA 67), the transmission spectrum between the two ports is measured. The cavity has a composite quality factor on the order of 370 and around 23 modes overlap at a given frequency~\cite{sol2022meta,sol2023reflectionless}. The system's scattering matrix is subunitary due to significant Ohmic losses on the metallic walls.
We randomly assign the available 152 1-bit programmable degrees of freedom of our programmable metasurface to 40 groups, coined macro-pixels, since our input vector's dimensions are $1\times 40$. All meta-atoms within a given macro-pixel are configured identically. We select {twenty} decorrelated frequencies within the metasurface's operation bandwidth and consider the transfer function intensity at these frequencies as outputs. The relation between inputs and outputs is hence non-linear due to the structural non-linearity~\cite{rabault2023tacit} and the readout non-linearity.
Our use of this microwave setup is distinct from previous works on over-the-air microwave analog computing with programmable metasurfaces in chaotic cavities. While, on the one hand, Ref.~\cite{del2018leveraging} also defined a mathematical operation in which the inputs were related to the metasurface configuration and the outputs to the transfer function, it aimed at implementing linear operations and hence sought to minimize the reverberation-induced structural non-linearity whereas we seek to maximize it in the present work. Refs.~\cite{sol2022meta,sol2023reflectionless}, on the other hand, considered the system's transfer function as the mathematical operation in order to implement  desired linear transfer functions for high-fidelity in situ programmable signal differentiation and routing; the inputs in Refs.~\cite{sol2022meta,sol2023reflectionless} were hence the incident wavefronts rather than the metasurface configuration.

\subsection*{S1-3. Optical system}
The data used for the optical experiment are from the published dataset in Rahmani et al. \cite{rahmani2020actor}. The optical system therein consists of a spatial light modulator, a scattering medium of a step-index multimode fiber of length $0.75$m and core diameter $50\mu$m with an aperture size 0.22, and a CMOS camera. The entire system is operating at the 532nm and 1 mW power continuous-wavelength light source corresponding to $\sim 1050$ modes of the fiber for one polarization. Using interferometry measurements \cite{loterie2015digital}, a transmission matrix of the optical system is obtained which allows mapping an input 2-dimensional optical field to its complex output field. The columns of the transmission matrix contain the response functions of the system for each of the modes of the fiber, allowing to faithfully calculate the optical responses of the system to arbitrary inputs.

\newpage
\section*{Section S2.  Comparison of different PNN training algorithms}

\begin{figure*}[t!]
	\centering\includegraphics[width=13.9cm]{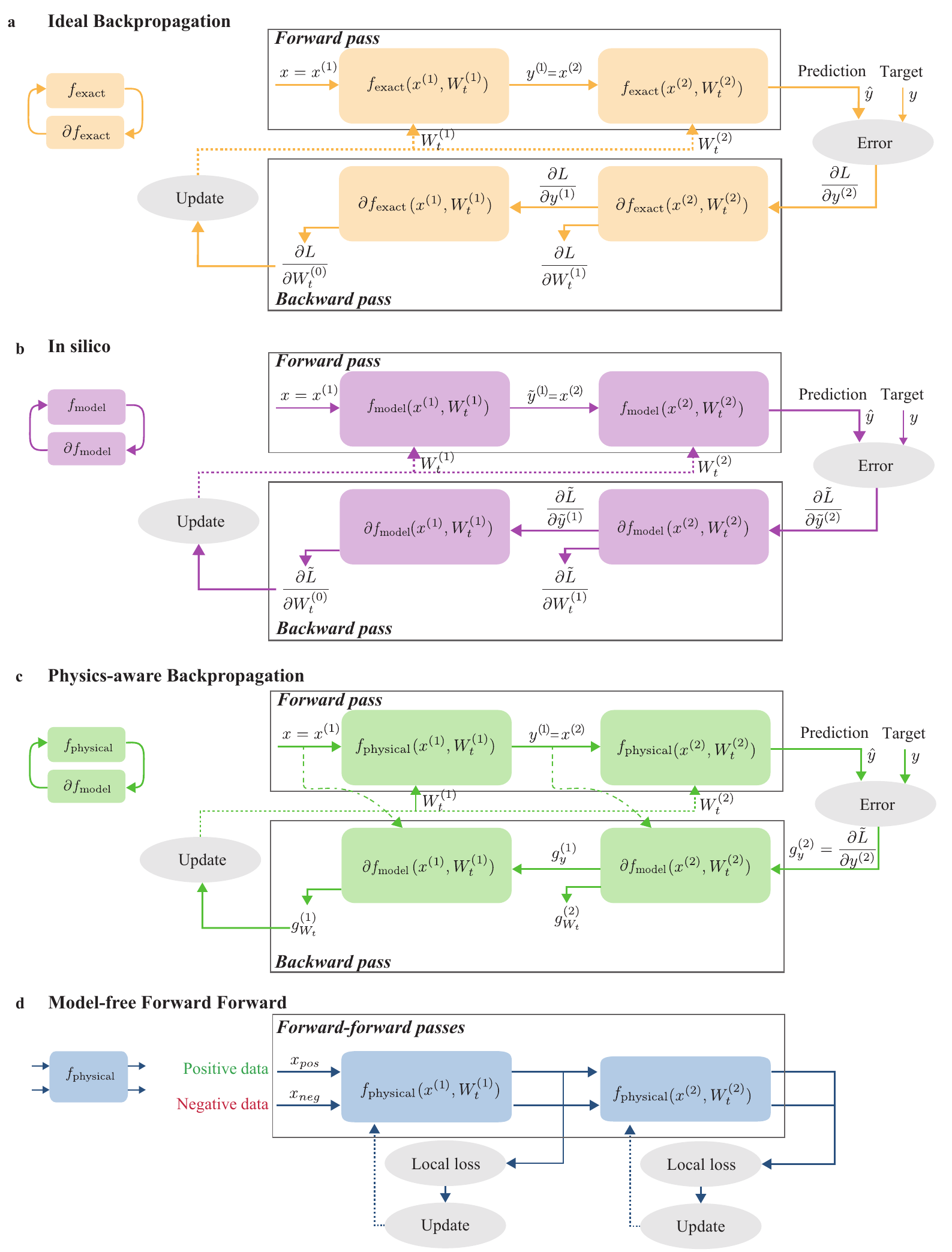}
	\caption{\textbf{ Comparison of different PNN training algorithms } \textbf{a}, Ideal BP. \textbf{b}, In-silico training. \textbf{c}, Physics-aware BP. \textbf{d}, MF-FF. }\label{Fig_s1}
\end{figure*}

In Figure \ref{Fig_s1}, a comprehensive comparison of different algorithms including ideal BP, in-silico, PA-BP, and MF-FF is depicted. 

All algorithms, except MF-FF, utilize backpropagation as the core technique and consist of four key steps: forward pass, error vector computation, backward pass, and parameter update. The key distinction among these algorithms lies in the choice of physical transformation, either $f_{physical}$ (physical transformation function) or $f_{model}$ (digital model transformation function), for the forward and backward passes. For example, the ideal BP uses the exact function of the system $f_{exact}$ for both the forward and backward passes. However, in-silico employs  $f_{model}$ for both the forward and backward passes. Physics-aware BP utilizes  $f_{physical}$ for the forward pass and  $f_{model}$ for the backward pass.

In contrast to the aforementioned methods for training PNN,   MF-FF employs two forward passes without employing any backpropagation through layers. The original paper was recently proposed by Geoffrey Hinton \cite{hinton2022forward}. The concept involves substituting the traditional forward and backward passes of backpropagation with two forward passes that function in the same manner, but with contrasting objectives and operate on distinct datasets. The positive pass involves processing real data and adjusts the weights to increase the goodness in every hidden layer. On the other hand, the negative pass involves processing negative data and adjusting the weights to decrease the goodness of each hidden layer. In this paper, we use the balanced contrastive loss which is recently proposed in Ref.~\cite{lee2023symba} (Eq.~1 in the main paper). However, other forms of imbalance positive-negative losses are possible \cite{hinton2022forward}. 

To update the parameters of each layer in conventional fully-connected multilayer neural networks using the forward-forward training algorithm, it is necessary to have knowledge of the nonlinear unit, activation function. Although the forward-forward algorithm augmented by gradient-free optimization techniques can be used to update parameters in each layer, it may impose some limitations, such as slow convergence rates, getting stuck in local minima, and requiring heavy digital computations. 
\begin{algorithm}[H]\small
\begin{algorithmic}[MF-FF]
    \INPUT Training dataset $(x_{i},y_{i})$, $N_{exter}$ and $N_{inter}$ (number of external and internal training epochs, respectively),  $L$ (number of layers), $\theta$ (scale factor), and $PNN_{forward}$ (forward pass of PNN) . 
    \newline
    \OUTPUT Trainable parameters of matrix $W_l$ for $l^{th}$ layer
    \newline
    \STATE \textbf{Initialization} Parameters $W_l$ 
  \FOR{Epoch \,$\in \{1,2,3,...N_{exter}\}$}
      \STATE Creating  $x_{pos}$ and $x_{neg}$ from $(x_{i},y_{i})$
      \STATE $h_{pos}, x_{pos} \gets h_{neg}, x_{neg}$
      \STATE 
      \FOR {layer \,$\in \{1,2,3,...L\}$}
            
             \STATE $h_{pos}, h_{neg} \gets PNN_{forward}(h_{pos}), PNN_{forward}(h_{neg})$
             
             \STATE $h_{pos},h_{neg} \gets \textbf{TrainLayer}(h_{pos},h_{neg})$

      \ENDFOR
      \STATE

      \ENDFOR
      \STATE 
      
      \STATE \textbf{Func: TrainLayer $(h_{pos},h_{neg})$}
      
      \FOR{Epoch \,$\in \{1,2,3,...N_{inter}\}$}
      \STATE $h_{\text{pos}},h_{\text{neg}} \gets W_l h_{\text{pos}},W_l h_{\text{neg}}$
      
       \STATE $ L \gets log \bigg(1+exp \bigg({-\theta \big(\sum_{j}{ h_{\text{pos},j}^{(l)}}^2 -\sum_{j}{ h_{\text{neg},j}^{(l)}}^2 \big) \bigg)}\bigg)$
      \STATE Updating $W_l$  by minimizing  L
      \ENDFOR
    
\end{algorithmic}
        \caption{\,}
        \label{alg1} 
\end{algorithm}
To mitigate such limitations, we employed a simple deep PNN architecture as illustrated in {Fig. 1a} of the main paper and explained in detail in the main paper. This enables simple gradient-based update rules, owing to the linearity of the trainable components in the PNN, which involve linear matrix multiplications. Algorithm \ref{alg1} provides a summary of the proposed MF-FF training algorithm.
\newline

\section*{Section S3. Acoustics-PNN}

\begin{figure}
	\centering
	\includegraphics[width=\textwidth]{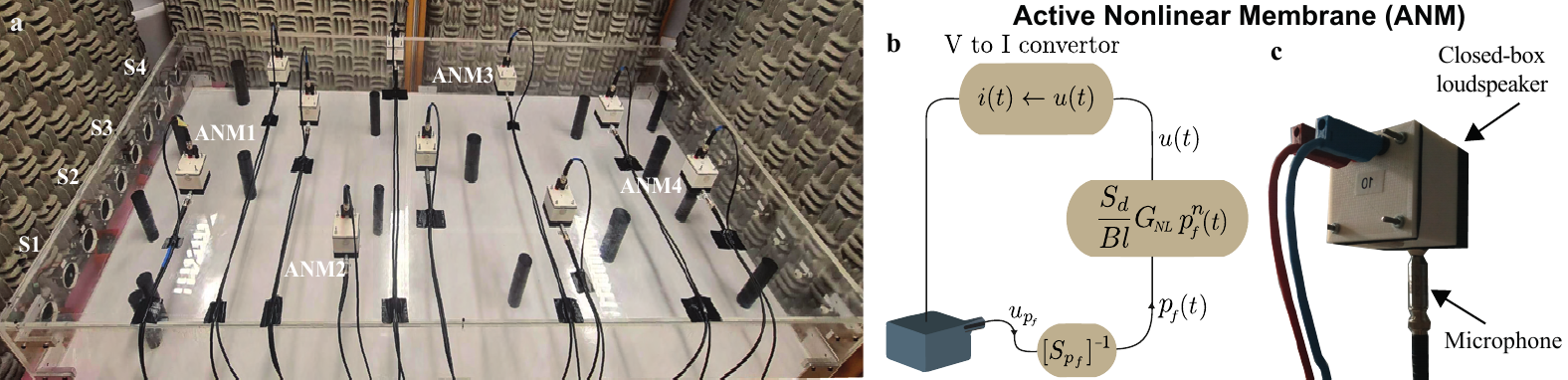}
	\caption{ \textbf{Active control scheme and set-up photography} \textbf{a}, Photography of the experimental set-up with the localization of the active nonlinear resonators. \textbf{b}, nonlinear active control scheme. \textbf{c}, Closed-up photography on the active nonlinear membrane (ANM). \label{Fig_s2}}
\end{figure}
\subsection*{ S3-1. Experimental setup}

The nonlinear acoustic transformer, shown in Fig. S2, consists of a large parallelepipedic cavity ($2\times 1\times 0. 2$ m) supporting 11 propagating modes below 500 Hz, instrumented with 10 source loudspeakers (Monacor SP6-8SQ) positioned vertically on the left side of the cavity, 10 nonlinear meta-scatterers on the upper rigid surface to increase the overall nonlinearity in the system, and 14 static cylindrical rigid scatterers scattered inside the cavity to increase the disorder and multiple scattering.

The training is performed using only 4 speaker sources out of the 10 available on the experimental apparatus. Each source delivers a different linear waveform with a content of 10 frequencies in the range $350-500$ Hz with randomly weighted amplitudes. The output of the system consists of 4 highly nonlinear pressure waveforms measured at the output of 4 of the actively controlled membranes (meta-scatterers at different locations.

Each meta-scatterer consists of an electrodynamic loudspeaker (Visator FRWS 5 SC), enclosed in a cavity of volume $V_c = 144$ m$^2$ and an ICP microphone (PCB 130F20, 1/4 inch) placed in front of the diaphragm. 
The non-linear active control described in Fig~\ref{Fig_s2}b consists of a real-time feedback loop that assigns a given current $i(t)$ to each NL resonator as a function of the measured front pressure and according to the following control law 
\begin{equation}
i(t) = \frac{Bl}{S_d} G_{NL}|p_f(t)|^{\alpha_{NL}},
\end{equation}
where $G_{NL}$ and $\alpha_{NL}$ are two tunable parameters that produce nonlinearity in the system, and $Bl$ and $S_d$ are the force factor and effective cross-sectional area of the speaker, respectively.

The optimal values for high nonlinearity and control stability for $G_{NL}$ and $\alpha_{NL}$ are reported in Table~\ref{tab:control_ac}.

\begin{table}[]
    \centering
    \begin{tabular}{|c|c|c|c|c|}
    \hline 
     & $G_{NL}$ & $\alpha_{NL}$ & $Bl$ ($Tm$) & $S_d$ ($m^3$) \\ 
    \hline 
    ANM1 & 5.7 & 1.6 & 1.1 & 0.0012 \\ 
    \hline 
    ANM2 & 5.5 & 1.7 & 1.0 & 0.0012 \\ 
    \hline 
    ANM3 & 5.3 & 1.5 & 1.0 & 0.0012 \\ 
    \hline 
    ANM4 & 5.6 & 1.4 & 1.1 & 0.0012 \\ 
    \hline 
    \end{tabular} 
\caption{\textbf{Control \& speaker parameters.}}
    \label{tab:control_ac}
\end{table}

The measurement is performed 10,000 times with different amplitude weightings for each run and each input. The set of input and output data is thus composed of two matrices with $40 \times 10 000$ elements.

\begin{figure}[t!]
	\centering
	\includegraphics[width=\textwidth]{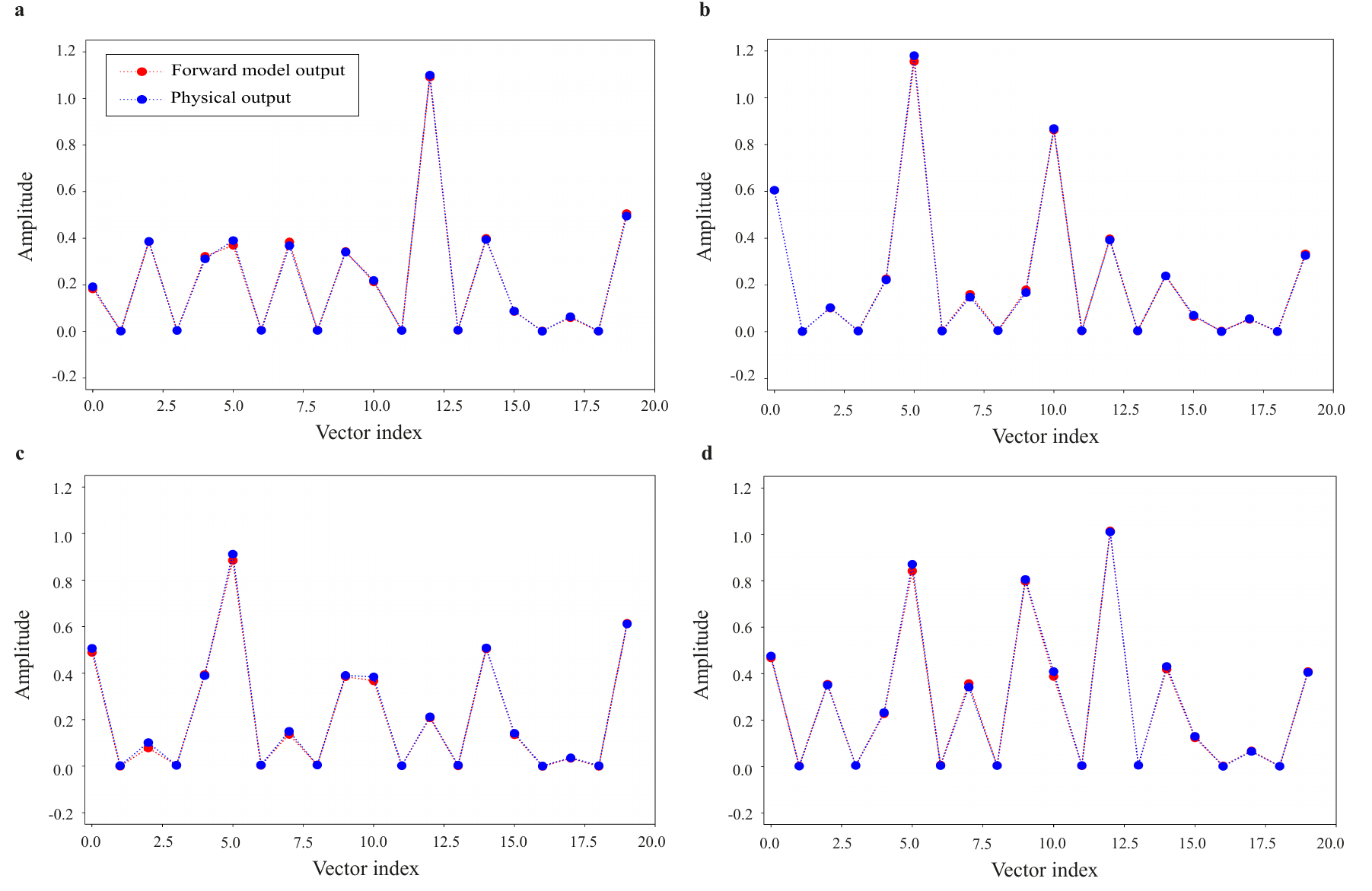}
	 \caption{ \textbf{Comparison of the forward model and physical acoustics-PNN for four randomly selected inputs from the test dataset.} \label{Fig_s3}}
\end{figure}

\subsection*{S3-2. Input-output transformation characterization}

To compare our results from the MF-FF method with other training methods such as ideal BP and in-silico training methods, we trained a digital neural network for the forward pass of the acoustic system (see {Fig. 2c} of the main paper).  
We collected training data by recording the outputs of the physical system for a selection of $N$ input
vectors, here N was $10^4$ and the input vectors were drawn from a uniform distribution. The data set was divided into training and validation sets and used to train the neural network. A fully-connected neural network was used to map the input-vector with the dimension of $1\times20$ to the output-vector with the dimension of $1\times 20$.   The numbers of hidden layers are three with  100, 200, and 100 neurons, respectively.  We used the Sigmoid Linear Unit (SiLU) activation function and Adam optimizer. After training for 200 epochs, the model achieved a mean squared error of $2.8\times 10^{-3}$ on the test dataset. In Fig. {\ref{Fig_s3}}, we randomly selected four samples from the testing dataset and compared the model's predictions to the ground truth physical output. This visual comparison allows us to assess how well the model predict the output of physical system.

The proposed acoustics-PNN consists of two layers with skip-connections (see {Fig. 2a} of the main paper) and was trained by MF-FF. We found that incorporating skip connections into the network was beneficial. This feature is particularly beneficial as the acoustic transformation lacks support for identity operation, making it susceptible to information loss. These connections, inspired by residual neural networks, were added to allow the network to act further as an ensemble of sub-networks \cite{veit2016residual,wright2022deep}, thus this contributes to the robustness of the architectures, allowing them to maintain high classification performance even in the presence of stochastic failures in any individual sub-component \cite{veit2016residual}.

In order to implement the results of in-silico (see {Fig. 2c} of the main paper), we added small noise (Gaussian noise with a mean of zero and standard deviation of 0.025) to the parameters of the forward model of the nonlinear acoustic system. As long as there is a small gap between the reality and the digital model of the physical system, the accuracy of inference will decrease, as illustrated in {Fig. 2c} of the main paper.    
\newpage
\section*{Section S4. Microwave-PNN}

\subsection*{S4-1. Experimental setup}

Our microwave system is shown in Fig.~3b of the main text and consists of an irregularly shaped electrically large metallic cavity ($0.385\ \mathrm{m} \times 0.422\ \mathrm{m} \times 0.405\ \mathrm{m}$; $0.0658\ \mathrm{m}^3$). Two waveguide-to-coax adapters (RA13PBZ012-BSMA-F) connect the cavity to two asymptotic scattering channels. Using a vector network analyzer (Rhode \& Schwarz ZVA 67), the transmission spectrum between the two ports is measured. The cavity has a composite quality factor on the order of 370 and around 23 modes overlap at a given frequency~\cite{sol2022meta,sol2023reflectionless}. The system's scattering matrix is subunitary due to significant Ohmic losses on the metallic walls.

\begin{figure}[h]
	\centering
	\includegraphics[width=1\textwidth]{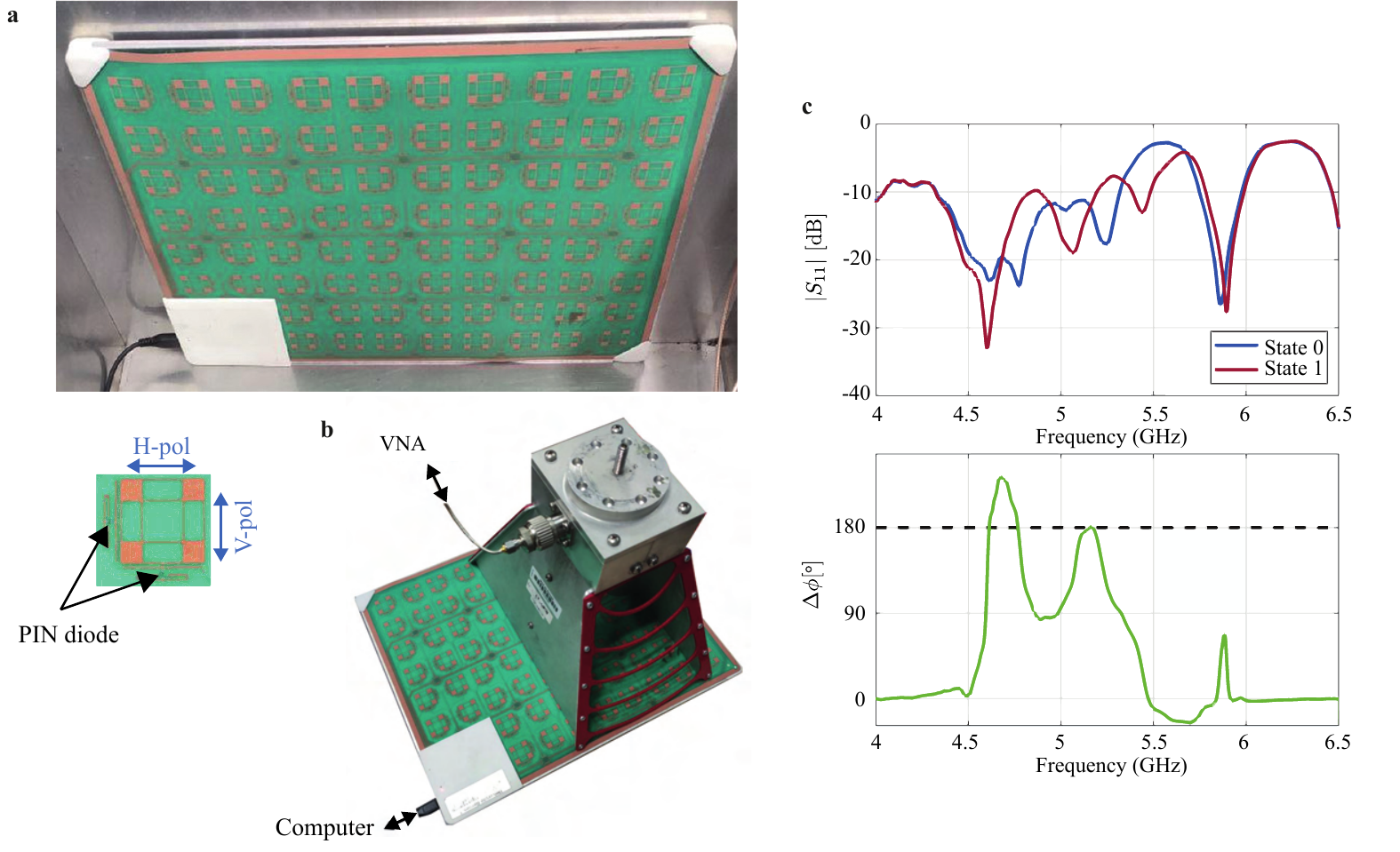}
	\caption{\textbf{Details about the programmable metasurface. } \textbf{a}, Close-up view.  \textbf{b}, Setup to characterize normal-incidence scattering properties of the metasurface. \textbf{c}, Normal-incidence characterization of the scattering response of the two different meta-atom configurations.\label{Fig_s4a}}
\end{figure}

The cavity's scattering properties are massively parametrized through a programmable metasurface. A close-up view of the programmable metasurface (purchased from Greenerwave) is shown in Fig.~\ref{Fig_s4a}a. The 2-bit programmable metasurface contains 76 meta-atoms that cover 8~\% of the cavity surface and efficiently modulate the field inside the cavity within a 400 MHz interval centered on 5.2 GHz. The two bits of control per meta-atom are assigned to the two orthogonal field polarizations (``H-Pol'' and ``V-Pol'' in the inset in Fig.~\ref{Fig_s4a}a). The working principle of the 1-bit control over each polarization follows that proposed in Ref.~\cite{kaina2014hybridized}. At the central working frequency (around 5.2~GHz), the metasurface can tune the reflected phase by roughly $\pi$, as seen in Fig.~\ref{Fig_s4a}c, implying that it can roughly mimic Neumann or Dirichlet boundary conditions.

We randomly assign the available 152 1-bit programmable degrees of freedom of our programmable metasurface to 40 groups, coined macro-pixels, since our input vector's dimensions are $1\times 40$. All meta-atoms within a given macro-pixel are configured identically. We select {twenty} decorrelated frequencies within the metasurface's operation bandwidth and consider the transfer function intensity at these frequencies as outputs. The relation between inputs and outputs is hence non-linear due to the structural non-linearity~\cite{rabault2023tacit} and the readout non-linearity.

Our use of this microwave setup is distinct from previous works on over-the-air microwave analog computing with programmable metasurfaces in chaotic cavities. While, on the one hand, Ref.~\cite{del2018leveraging} also defined a mathematical operation in which the inputs were related to the metasurface configuration and the outputs to the transfer function, it aimed at implementing linear operations and hence sought to minimize the reverberation-induced structural non-linearity whereas we seek to maximize it in the present work. Refs.~\cite{sol2022meta,sol2023reflectionless}, on the other hand, considered the system's transfer function as the mathematical operation in order to implement  desired linear transfer functions for high-fidelity in situ programmable signal differentiation and routing; the inputs in Refs.~\cite{sol2022meta,sol2023reflectionless} were hence the incident wavefronts rather than the metasurface configuration.

Finally, we now detail two additional characterizations of our setup.

\begin{figure}[h]
	\centering
	\includegraphics[width=1\textwidth]{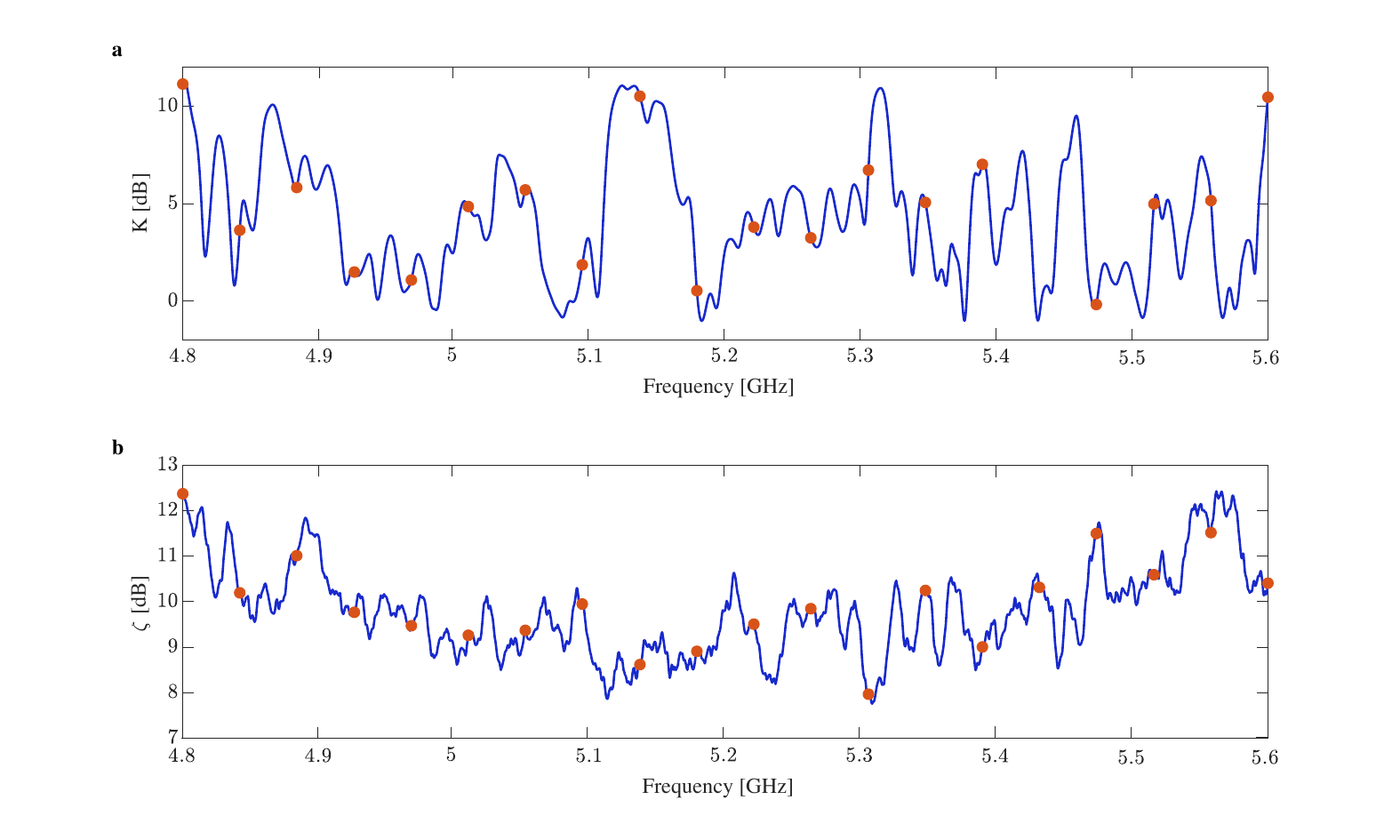}
	\caption{\textbf{Additional characterization of the microwave setup. } \textbf{a}, Field stirring efficiency.  \textbf{b}, Linearity metric.\label{Fig_s4}}
\end{figure}

First, in Fig.~\ref{Fig_s4}a, we quantify the control of the programmable metasurface over the transmission spectrum between our two antennas in terms of the $K$-factor that is commonly used in electromagnetic compatibility and wireless communications. The $K$-factor is defined at each frequency as the power ratio of the unstirred to the stirred field components: $K(f) = |\mu(f)|^2 / 2[\sigma(f)]^2$, where $\mu(f)$ and $\sigma(f)$ are the mean and standard deviations of an ensemble of transmission spectra corresponding to random metasurface configurations. The lower the K-factor is, the more efficiently the metasurface stirrs the field, i.e., the more control it has over the transmission between the two antennas. We have highlighted in Fig.~\ref{Fig_s4}a the 20 frequency (regularly spaced) that we selected for our analysis in the main text. The stirring efficiency is seen to be quite frequency dependent, since it is determined both by the metasurface's frequency response and by the frequency-dependent scattering properties of the cavity.

Second, in Fig.~\ref{Fig_s4}b, we quantify the linearity of the complex-valued transmission spectrum's dependence on the metasurface configuration. Recall that in the main text we only work with the intensity rather than the complex-valued fields analyzed here, which adds an additional readout non-linearity. Here, we seek to only quantify the structural non-linearity. We use the linearity metric $\zeta$ introduced in Ref.~\cite{rabault2023tacit}. To compute this metric at a given frequency, we begin by fitting the best possible linear model, $h = h_0 + t^Tc$, to the data set of random metasurface configurations $c$ and the corresponding transmission coefficients $h$. Then, we compute the linear model's error on unseen data, $\Delta h$, and evaluate an SNR-like metric that treats this prediction error like noise in the usual SNR definition: $\zeta = SD_i(h_i) / SD_i(\Delta h_i)$. The higher the value of the linearity metric $\zeta$ is, the more linear is the relation between the transmission spectrum and the metasurface configuration. We observe that the value of $\zeta$ is lowest in the middle of the metasurface's frequency interval of operation, and that it is always in the vicinity of 10~dB. These values evidence that there are significant non-linear effects in the mapping from metasurface configuration to transmission spectrum (which we refer to as structural non-linearity) that were shown in Ref.~\cite{rabault2023tacit} to originate from two factors: (i) mutual coupling due to proximity between close-by meta-atoms; and (ii) reverberation-induced long-range coupling between all meta-atoms. Again, we have highlighted in Fig.~\ref{Fig_s4}b the 20 frequencies that we selected for our analysis in the main text.

\subsection*{S4-2. Input-output transformation characterization}

\begin{figure}[h]
	\centering
	\includegraphics[width=\textwidth]{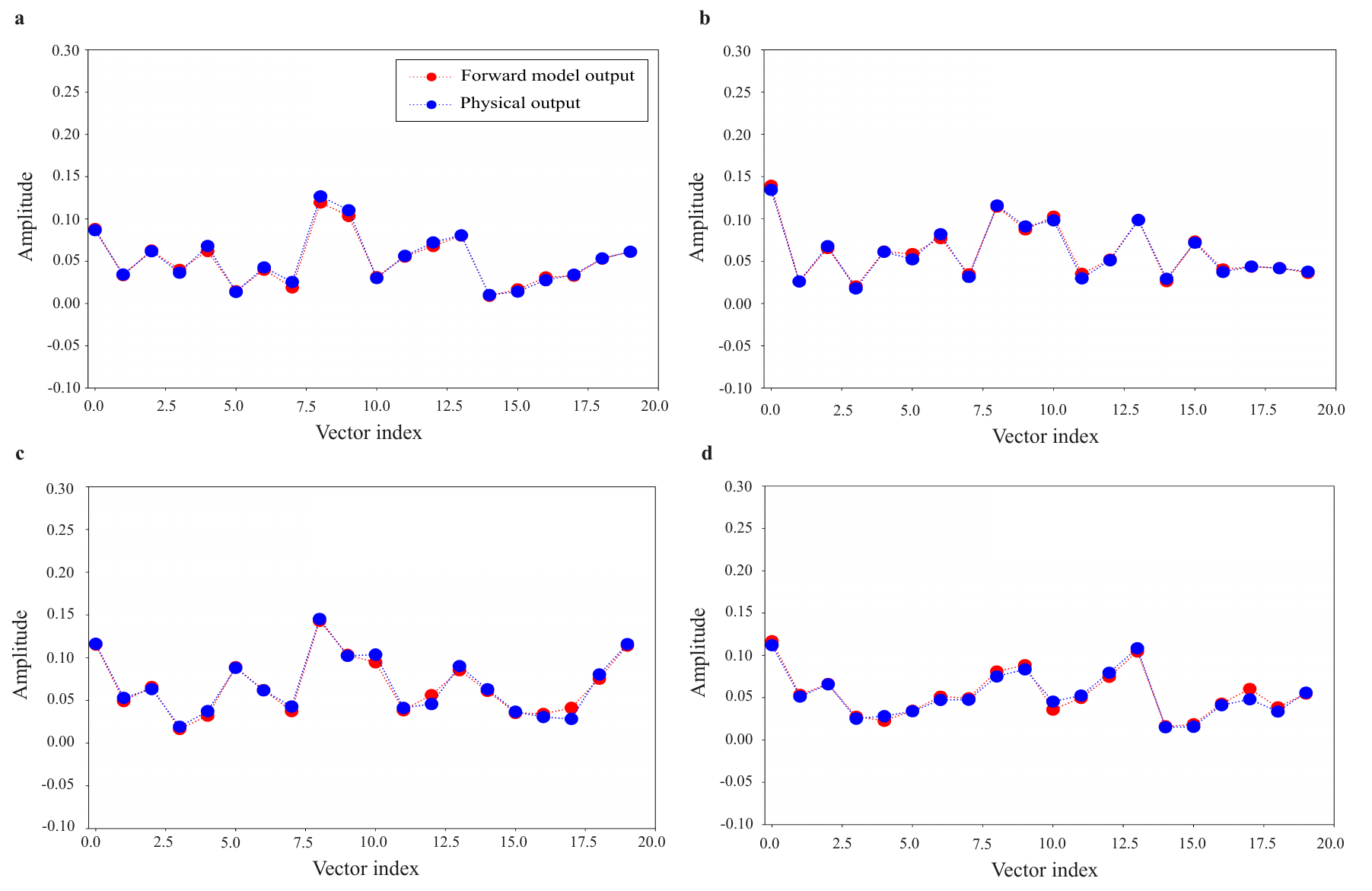}
	\caption{\textbf{Comparison of the forward model and physical microwave-PNN for four randomly selected inputs from the test dataset.} \label{Fig_5}}
\end{figure}

Similar to the procedure outlined for the acoustics-PNN, we trained a digital neural network to serve as a surrogate forward model that approximates the forward pass of our microwave system
(see Figure~3a of the main paper). We obtained training data by recording the physical system's outputs for a chosen set of $N=5\times10^5$ random input vectors. The input vectors were randomly selected from a binary uniform distribution (since we have 1-bit control over each programmable degree of freedom). We use a fully-connected neural network to map the input vector with dimensions of $1 \times 40$
to the output vector with dimensions of $1 \times 20$. It consists of nine hidden layers with 100, 200, 400, 800, 800, 400, 200, 100, and 20 neurons, respectively. We used the Sigmoid Linear Unit (SiLU) activation function and Adam. We used layer normalization and dropout in order to prevent overfitting.  After training for 200 epochs, the model achieved a mean squared error of $2.4\times 10^{-5}$ on the test dataset. In Figure {~\ref{Fig_5}}, we randomly selected four samples from the testing dataset and compared the model's predictions to the ground truth, physical output.

The proposed microwave-PNN consists of three layers with skip-connections (see Fig. 3a of the main paper) and was trained by MF-FF.

\newpage
\section*{Section S5. Optics-PNN}

\subsection*{S5.1 Experimental setup}

The experimental system from which the optical dataset is obtained is explained in detail in the previous work \cite{rahmani2020actor}.

\subsection*{S5-2. Input-output transformation characterization}
\begin{figure}[t!]
	\centering
	\includegraphics[width=13.5cm]{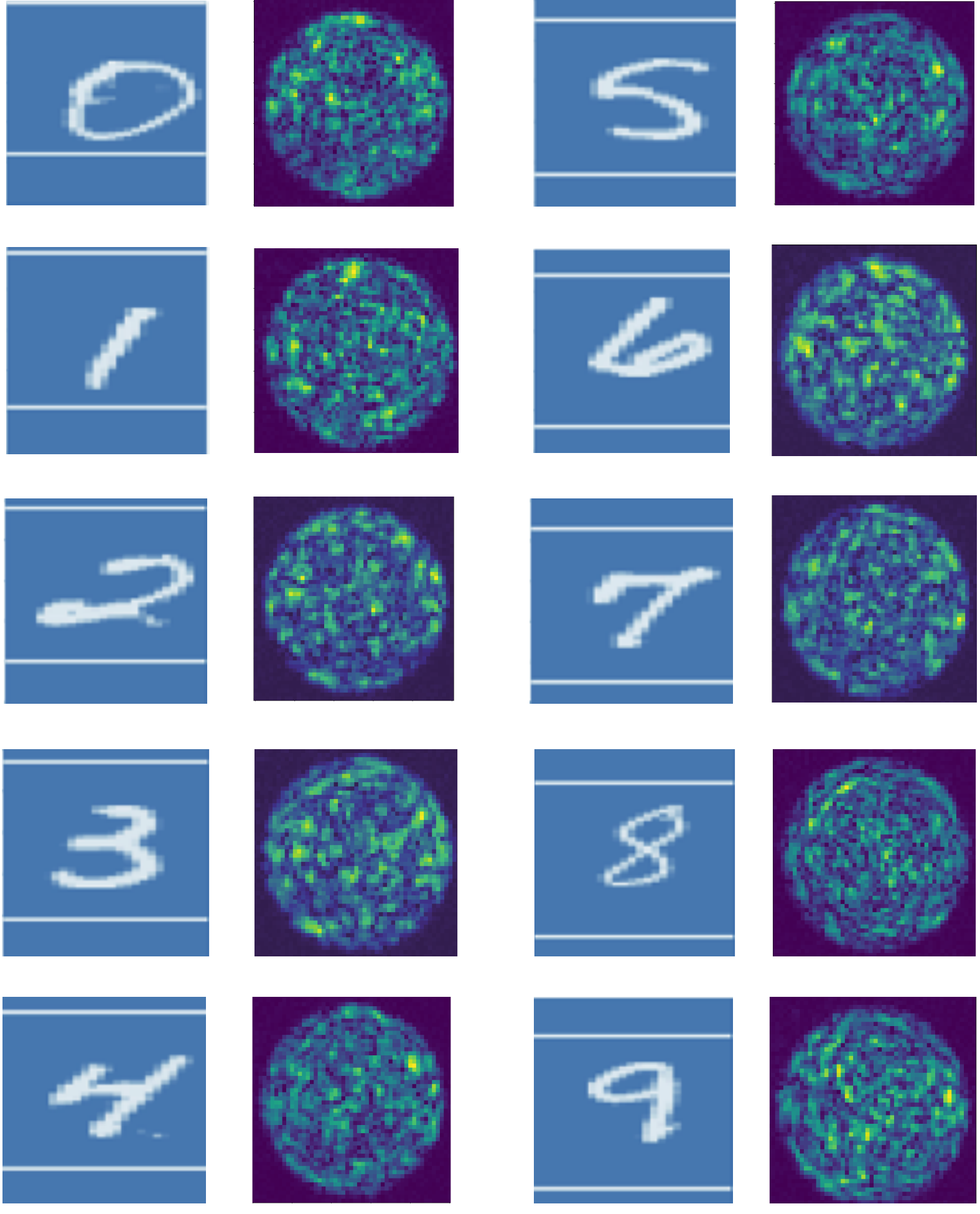}
	\caption{\textbf{ Transformation of optics-PNN for ten randomly selected digits from the Mnist dataset.} \label{Fig_s6}}
\end{figure}

The MMF transformation consists of three steps: first, a Fourier transform is applied to the input image; second, the result is multiplied by a transmission matrix; and third, an inverse Fourier transform is applied to the product. Figure {\ref{Fig_s6}} displays the transformation of optics-PNN for ten randomly selected digits from the Mnist dataset. 

The proposed optics-PNN consists of two layers (see Fig. 4a
of the main paper) and was trained by MF-FF. The dimension of matrix multiplications is $51\times51 $ and $676\times676 $ for the Vowel and Mnist datasets, respectively.

\newpage
\newpage
\pagebreak
\section*{Section S6. Energy consumption and computing rate analysis}

Here we discuss the energy cost of the physical neural network proposed in this manuscript. We analyze the energy cost of a layer of the physical neural network for the inference and training stages separately. The FF training is also particularly suitable for breaking down the energy estimation per layer.

\begin{table}[]
    \centering
\begin{tabularx}{\textwidth}{ |X||X|X| }
    \hline 
     & Energy Consumption &  Details and/or  Ref   \\ 
    \hline 
    $E_{read}$ & 1 pJ/bit for off-chip memory, and
0.3 pJ/bit for SRAM &  \cite{jin2020user,anderson2023optical}\\ 
    \hline 
    $E_{DAC}$  & 10 pJ per 5-bit sample at 10 GHz & This is achievable
with 100 mW at 30.1dB SFDR \cite{anderson2023optical}\\ 
    \hline 
    $E_{mod}$  & $45\times$ $r_t$ pJ /MAC  & by considering the 1mW continuous laser beam.   $r_t$ is the refresh time corresponding to the refresh time of SLM or DMD.  
\cite{rahmani2020actor}\\ 
    \hline 
    $E_{SLM}$  & $15 nJ $ /MAC &  Assuming 2W for operation
of a $ 1920 \times 1152$ SLM, working at  60 Hz   \cite{anderson2023optical}\\
    \hline 
    $E_{DMD}$  &$ 94 pJ  $ /MAC  & For Texas Instruments DLP9500  DMD model (with a $1920 \times 1080$ resolution and  23148 patterns per second ) \cite{oguz2022programming}\\ 
    \hline 
    $E_{ADC}$ &   3.17 pJ per 7-bit sample & \cite{anderson2023optical}\\ 
    \hline 
    $E_{read out}$ &   1.4 nJ /MAC
    
    & For LUX 1310 camera \cite{oguz2022programming}\\ 
    \hline 
\end{tabularx} 

\caption{\textbf{The energy budget breakdown of the  optical computing system.}}
   \label{optical}
\end{table}
At the inference stage, the energy consumption can be broken down into the energy cost of the data loading into the physical system, a feedforward pass of the wave fields in the system, and a readout phase into the digital system \cite{anderson2023optical, pai2022experimentally}. Therefore,

\begin{equation}
E_{inference}= E_{load} + E_{feedforward} + E_{readout}
\end{equation}

The $E_{load}$ consists of the energy required to read from the memory, digital to analog (DAC) conversion as well as the modulation energy of the wave modulator ($E_{load}$ = $E_{read}$ + $E_{DAC}$ + $E_{mod}$). Similarly, the $E_{readout}$ is comprised of a wave amplification, analog to digital conversion (ADC), and writing to memory (or energy consumption of camera). The $E_{feedforward}$ consists of the energy of the waves in the system including the wave sources in addition to maintaining any active components in the system including SLM or DMD. The details of energy consumption are listed in Table S2 for the proposed optical system. 

At the training phase, the energy consumption consists of the energy at the inference $E_{inference}$ plus the energy consumption of a single gradient update of the learnable parameters.  The $E_{gradient}$ consists of a \textit{digital} gradient update to the learnable parameters of the physical network as well as an additional $E_{load}$ to implement a \textit{physical} update to the parameters. Therefore,

\begin{equation}
E_{training}= 2E_{load} + E_{feedforward} + E_{readout} + E_{gradient}
\end{equation}

In the above analysis, the energy contributions of the $E_{load}, E_{readout}$ and $E_{feedforward}$ scale linearly with the number of the data points $N$ (image's pixels, vowel's frequencies, etc.) loaded/read out of the system (we assume the system has the same number of inputs and outputs) while $E_{gradient}$ having an $N^2$ scaling contribution. 

The energy consumption for the training/inference of a fully digital neural network is measured in the total number of digital floating-point operations per second (Flops) or the total number of multiply–accumulate (MAC) operations of a given vector-matrix product. In our proposed FF-model, the digital energy cost of a forward pass through the linear layer requires quadratic $N^2$ MAC operations while a PNN's energy scales linearly, i.e. $N(E_{load} + E_{readout})$. While the cost of $E_{feedforward}$ is virtually free, $E_{load}$, $E_{readout}$ are the leading contributors to the total energy consumption of a PNN. However, in large models, these extra costs amortize the per-MAC cost of the PNN compared to the constant per-MAC cost of a digital model.

While PNNs hold great potential for decreasing the energy cost of the computation, the computing rate of a PNN could be unpleasantly slow. Wave modulators are usually bottlenecking this information processing speed. Spatial light modulators (SLMs) and digital micromirror devices (DMDs)
are currently the major optical modulators with speeds on the orders of Hzs and kHzs. For inference, the effective computation rate is limited by the rate of the input data fed to the system ($\sim GHz$) whereas for training, the rate is limited to the refresh rates of modulators containing the learnable weights of the system. Switching from an LC SLM to a DMD could improve the inference speed by nearly 3 orders of magnitude,
reaching up to 23000 frames/second. Although the slow computation rate of the training is amortized by the fast rate of the inference (training is done only once), improvements to electronics of the modulators could further improve the training efficiency of PNNs in the future.

\end{document}